\documentclass{article}
\usepackage{ijcai26}
\usepackage{times}
\usepackage{soul}
\usepackage{url}
\usepackage[hidelinks]{hyperref}
\usepackage[utf8]{inputenc}
\usepackage[small]{caption}
\usepackage{graphicx}
\usepackage{amsmath}
\usepackage{amsthm}
\usepackage{booktabs}
\usepackage{algorithm}
\usepackage{algorithmic}
\usepackage[switch]{lineno}
\usepackage{amsfonts}
\usepackage{subcaption}
\usepackage{xcolor}

\title{Active Diffusion-Based Inference for Ill-Posed Inverse Problems under Incomplete Priors}

\author{
Jitao Xu$^1$
\and
Nobuo Sato$^2$\and
Yaohang Li$^{1}$\\
\affiliations
$^1$Department of Computer Science, Old Dominion University, Norfolk, Virginia 23529, USA\\
$^2$Jefferson Lab, Newport News, Virginia 23606, USA\\
\emails
jxu004@odu.edu,
nsato@jlab.org,
yaohang@cs.odu.edu
}
\begin{document}

\maketitle

\begin{abstract}
    Many scientific and engineering applications require estimating unknown parameters from experimentally observable data -- an inverse problem that is inherently challenging due to nonlinearity, noise, and ill-posedness.
    In this paper, we propose an active diffusion-based inverse problem solver. 
    A DM is trained to learn the mapping between the parameter space and the observable space.
    By iteratively detecting and correcting model misspecification through posterior uncertainty, the method discovers and learns the correct region of parameter space, even when initial training bounds exclude the true parameters. 
    This provides a principled, Bayesian justification for adaptive domain augmentation and ensures robust inference for inverse problems under incomplete prior knowledge.
    We demonstrate the effectiveness of our inverse solver for a toy inverse problem with infinite solutions, and for the parameterization of the quantum correlation functions to event observables in a Quantum Chromodynamics analysis of nucleon structure.
\end{abstract}

\section{Introduction}
Recovering unknown model parameters from measured observables is a fundamental inverse problem that arises across a wide range of scientific and engineering disciplines, including medical imaging~\cite{KruseSosonkinaBinHossenLi}, remote sensing ~\cite{rodgers2000inverse,yuan2020deep}, non-destructive inspection~\cite{nondestructiveinspection}, and nuclear physics~\cite{VAIMCFF}.
In these applications, the forward model mapping from parameters to observables is well-posed: given a set of parameters, the corresponding observables can be uniquely determined, although carrying out the forward computation can be computationally costly, requiring solving differential equations or applying linear algebraic operators.
In contrast, the inverse problem, mapping from observable space back to parameter space, is often ambiguous, unstable, and sensitive to noise. 
As a result, inverse problems are typically ill-posed, ill-conditioned, with significantly greater computational challenges~\cite{tarantola2005inverse}.

Recent advances in machine learning have introduced generative AI models as powerful alternatives to solve inverse problems.
Unlike deterministic regression-based approaches, these generative models, such as Invertible Neural Networks (INN)~\cite{ardizzone2018analyzing}, Variational Autoencoder Inverse Mapper (VAIM)~\cite{vaim}, Normalizing Flows (NF)~\cite{dinh2014nice,rezende2015variational}, and Diffusion Models (DM)~\cite{sohl2015deep,song2019generative,ho2020denoising}, learn probability distributions over parameters and observables, enabling inference through sampling.
By explicitly modeling uncertainty and generating posterior solution distribution from sampling a latent space, generative AI models provide a natural mechanism for addressing the ill-posedness issue that is intrinsic to inverse problems~\cite{song2020score,kawar2022denoising,chung2022diffusion}.

However, the generative AI-based inverse solvers implicitly assume that the admissible parameter space is known in advance. 
In practice, although sometimes the approximate guesses are available, precise bounds on the true parameters are often unknown, particularly in scientific discovery problems. 
When the true parameters lie outside the training domain, learned inverse solvers have to extrapolate unpredictably and produce misleading confidence.
Particularly, the generative models can fit complex distributions but do not inherently signal when they are operating beyond their training support.

In this paper, we propose an active diffusion-based inverse solver without precise prior knowledge of the parameter domain.
Here, DMs are trained to learn the mapping between parameter space and observable space and can be used to do inference by sampling from the conditional distribution of parameters when observables are given.
Rather than relying on the fixed parameter bounds, active learning method ~\cite{settles2009active} is adopted to iteratively detect model misspecification through posterior distribution analysis and actively augments the training domain when necessary.
By generating new forward simulations in regions indicated by the model's inference uncertainty, our method progressively discovers and learns the relevant region of parameter space with respect to the true parameters, even when they are outside the initial bounds.
A principled Bayesian justification is provided for adaptive domain augmentation.
We demonstrate the effectiveness of the proposed method on a toy inverse problem with infinitely many solutions, as well as on a realistic application in nuclear physics involving the parameterization of quantum correlation functions and their mapping to event-level observables in a Quantum Chromodynamics analysis of nucleon structure.

\section{Related Work}
Inverse problems are fundamentally ill-posed with non-unique solutions and unstable, which substantially increases their computational complexity.
Classical numerical approaches thus rely on incorporating additional assumptions or regularization, such as sparsity~\cite{foucart13} or low dimensionality~\cite{ijcai2017-468}, to constrain the solution space.
Beyond these generic assumptions, many inverse problem solvers are highly application specific, which depend on detailed domain knowledge.
Comprehensive surveys of classical approaches to ill-posed inverse problems across various scientific disciplines can be found in the literature~\cite{IPSurvey}.

The emergence of data-driven methods and deep learning has significantly expanded the set of available methods for solving inverse problems~\cite{arridge2019}.
Deep neural networks have been used to learn the mappings between observables and parameters directly from data, often outperforming traditional numerical methods in not only accuracy, but also computational efficiency.
Prior work has explored incorporating deep neural networks as learned regularizers~\cite{Li_2020} and extracting prior knowledge~\cite{Alder2017}.

More recently, generative modeling has provided a probabilistic framework for inverse problems by explicitly learning distributions over parameters and observables. 
Mixture density networks (MDNs)~\cite{Bishop94mixturedensity} represent one of the earliest generative modeling approaches for probabilistic inversion by modeling the conditional posterior as a weighted mixture of parametric distributions, enabling the representation of multimodal solutions but often relying on restrictive distributional assumptions.
Invertible neural networks (INN)~\cite{ardizzone2018analyzing} enable exact likelihood evaluation and uncertainty quantification through bijective mappings, but their architectural constraints can limit expressivity. 
Variational autoencoder–based approaches (VAIM)~\cite{vaim} introduce latent variables to capture multimodal posterior distributions, though approximation errors in the variational objective may affect inference quality. 
Normalizing flow (NF) models~\cite{dinh2014nice,rezende2015variational} offer expressive density estimation with tractable likelihoods, but similarly require strict invertibility and Jacobian constraints.
Diffusion models (DM)~\cite{sohl2015deep,song2019generative,ho2020denoising} have emerged as a highly expressive alternative, capable of modeling complex high-dimensional distributions through iterative denoising without requiring explicit invertibility.

Several recent studies have applied DMs to inverse problems by conditioning the generative process on observed data, demonstrating strong performance in image reconstruction and scientific inference tasks.~\cite{ictai2025}
However, most existing approaches assume that the admissible parameter domain is known in advance and fixed during training. 
When this assumption is violated, DMs, as expressive generative models, extrapolate beyond their training support without reliably signaling model misspecification. 
Addressing this limitation remains an open challenge.

Building on the generative approaches, this work introduces an active diffusion-based framework for solving inverse problems, which adaptively expands the parameter domain during training.
By detecting inconsistencies through posterior analysis and/or ensemble disagreement, the method iteratively augments the training support. 
This distinguishes our approach from prior diffusion-based inverse solvers, providing an effective mechanism for addressing incomplete prior knowledge in inverse problems.

\section{Methods}
We consider an inverse problem where a set of unknown physical parameters $\theta \in \mathbb{R}^d$ to be inferred from experimentally measured observables $y \in \mathbb{R}^m$ with a forward model 
\[
    \mathcal{F} : \theta \mapsto y.
\]
For a given observable $y^\star$, our goal is to estimate \(\theta^\star\) satisfying
\[
    y^\star = \mathcal{F}(\theta^\star).
\]

While the forward model $\mathcal{F}$ may be known, the inverse $\mathcal{F}^{-1}$ is often analytically intractable and is probably not unique. Since the admissible parameter domain \(\Theta\) is unknown \emph{a priori}, training a generative surrogate over a fixed range may exclude the true parameters. 
To overcome this limitation, we propose an active diffusion-based framework that adaptively expands and refines the support of the training distribution until it captures the region of parameter space consistent with the observed data.
We begin by introducing posterior variance as an indicator of domain expansion, and then present a more practically reliable indicator based on ensemble inference.

\subsection{Diffusion Model for Forward and Inverse Mapping}

We train a conditional DM \(p_\phi(\theta_t \mid y, t)\) to approximate the data
distribution \(p(\theta)\) and the conditional distribution \(p(\theta \mid y)\),
where \(t\) denotes the diffusion timestep and \(\phi\) the model parameters.
During training, parameters are drawn from an initial guessed domain \(\Theta_0\), and synthetic
observables are generated using the forward model:
\[
    y = \mathcal{F}(\theta).
\]
The DM is trained to represent the joint distribution
\[
    p(\theta, y) = p(\theta) p(y \mid \theta),
\]
with \(y\) serving as conditioning information during the reverse diffusion process. Given a fixed
observable \(y^\star\), the reverse-time sampling process approximates the posterior over parameters:
\[
    \theta \sim p_\phi(\theta \mid y^\star).
\]
We denote by \(\hat{\theta}\) the posterior mean and by \(\Sigma_\theta\) its covariance.

% \subsection{Parameter-Space Uncertainty Estimation as a Diagnostic}

% Because the initial domain \(\Theta_0\) may not contain the true parameters, the diffusion model may
% be forced to extrapolate outside its training support. In such situations, the inferred posterior
% variance becomes inflated. Formally, if \(\theta^\star \notin \Theta_0\), then after training,
% \[
%     \mathrm{Tr}(\Sigma_\theta) \to \text{large}.
% \]
% Hence, the posterior variance acts as an estimator of the model misspecification error:
% \[
%     \mathcal{E}_{\mathrm{miss}} \approx \mathrm{Tr}(\Sigma_\theta).
% \]
\subsection{Parameter-Space Uncertainty Estimation as a Diagnostic}

Because the initial domain \(\Theta_0\) may not contain the true parameters, the DM may
be forced to extrapolate outside its training support. In such situations, the inferred posterior
variance becomes inflated. Formally, if \(\theta^\star \notin \Theta_0\), then after training,
\[
    \mathrm{Tr}(\Sigma_\theta) \to \text{large}.
\]
Hence, the posterior variance acts as an estimator of the model misspecification error:
\[
    \mathcal{E}_{\mathrm{miss}} \approx \mathrm{Tr}(\Sigma_\theta).
\]

\subsection{Active Learning Loop}
\label{activelearningloop}

To correct for insufficient training support, we introduce an active-learning procedure that expands
the parameter domain adaptively. The algorithm is described as follows.

\paragraph{Step 1: Initial guess.}
Select an initial parameter domain \(\Theta_0\) based on a physical guess and generate a dataset on the initial training region
\[
    \mathcal{D}_0 = \{ (\theta_i, \mathcal{F}(\theta_i)) : \theta_i \in \Theta_0 \}.
\]
Train the DM on \(\mathcal{D}_0\).

\paragraph{Step 2: Inference on target data.}
Given the control observable \(y^\star\), perform reverse diffusion to obtain inferred parameter samples
\[
    \tilde{\theta} \sim p_\phi(\tilde{\theta} \mid y^\star),
\]
and compute \(\Sigma_{\tilde{\theta}}\).

\paragraph{Step 3: Uncertainty-triggered domain update.}
If the uncertainty is small, the inferred parameters are deemed reliable.  
If the uncertainty is large, i.e.,
\[
    \mathrm{Tr}(\Sigma_{\tilde{\theta}}) > \tau,
\]
for some threshold \(\tau\), we treat the inferred samples $\tilde{\theta}$ as informative proposal
points indicating missing training support. 
$\tau$ encodes the acceptable level of posterior uncertainty for a specific inverse problem and should be chosen with respect to observation noise and desired accuracy.
Specifically, we define an expanded domain
\[
    \Theta_{k+1} = \mathrm{Expand}(\Theta_k, \tilde{\theta}),
\]
where the expansion operator enlarges \(\Theta_k\) to include $\tilde{\theta}$.
New synthetic samples are then generated:
\[
    \mathcal{D}_{k+1} =
    \{ (\theta_j, \mathcal{F}(\theta_j)) : \theta_j \in \Theta_{k+1} \}.
\]
The DM is fine-tuned on the augmented dataset $\mathcal{D}_{k+1}$.

\paragraph{Step 4: Iteration.}
Steps 2–3 are repeated until
\[
    \mathrm{Tr}(\Sigma_{\tilde{\theta}}) \le \tau,
\]
indicating that the posterior has contracted to a well-supported region containing the true
parameter \(\theta^\star\).
% \begin{figure}[t]
%     \centering
%     \includegraphics[width=\columnwidth]{Figures/WorkflowImage.png}
%     \caption{Overview of the active diffusion-based inference framework. Left (Training Phase): parameters are sampled from the prior domain, passed through the forward model to generate training data, and used to train the DM. Right (Inference Phase): given a target observable $y^*$, the trained model performs inference; if the estimated uncertainty exceeds the threshold $\tau$, the domain is adaptively expanded and the model is retrained until convergence.}
%     \label{fig:workflow}
% \end{figure}
\begin{figure}[t]
    \centering
    \includegraphics[width=\columnwidth]{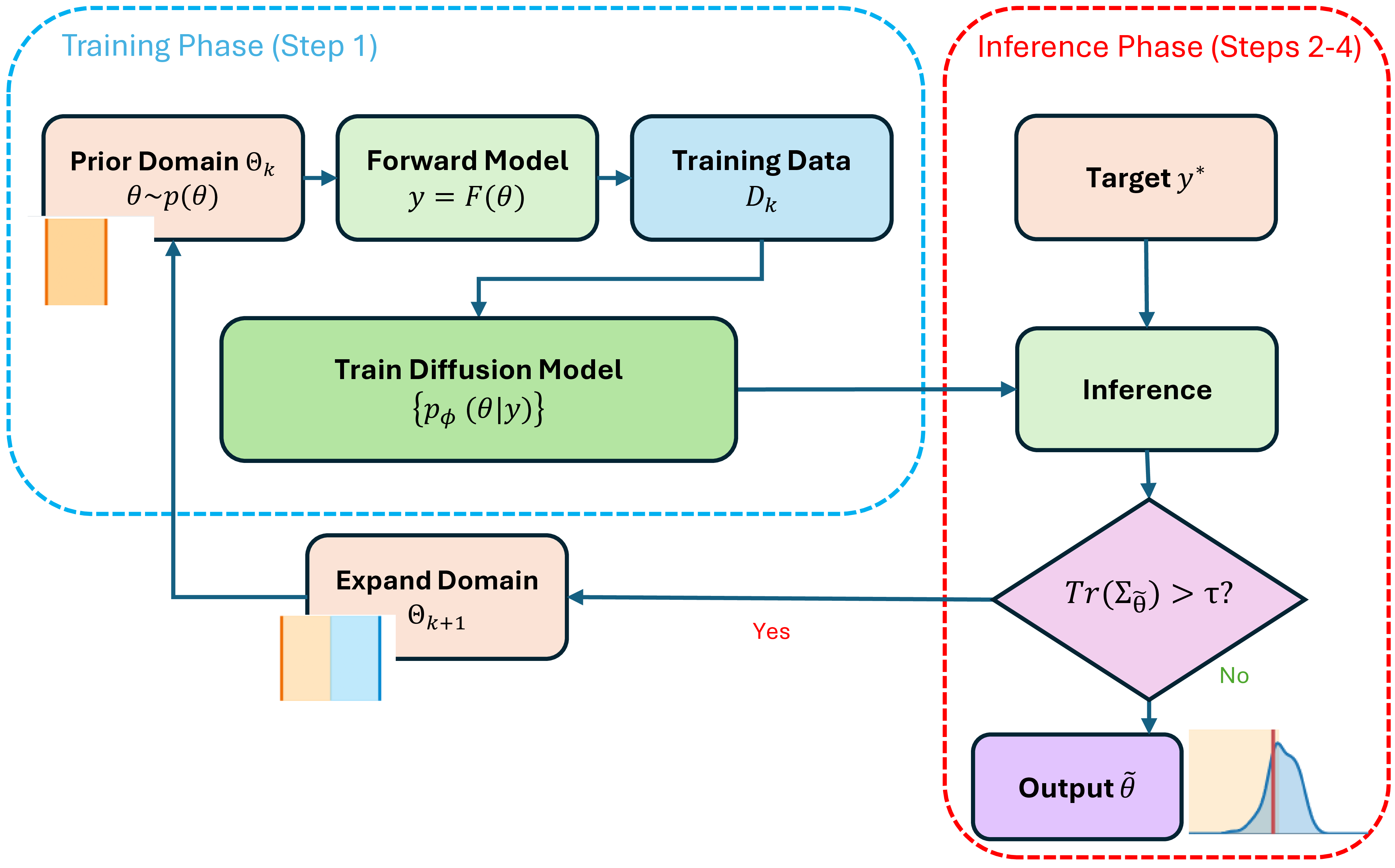}
    \caption{Overview of the active diffusion-based inference framework. Left (Training Phase): parameters are sampled from the prior domain, passed through the forward model to generate training data, and used to train the DM. Right (Inference Phase): given a target observable $y^*$, the trained model performs inference; if the estimated uncertainty exceeds the threshold $\tau$, the domain is adaptively expanded and the model is retrained until convergence.}
    \label{fig:workflow}
\end{figure}
Figure~\ref{fig:workflow} illustrates the overall active diffusion-based inverse problem solver workflow. The training phase (Step 1) samples parameters from the current domain, generates observables through the forward model, and trains the DM. The inference phase (Steps 2--4) performs inference on the target observables, evaluates uncertainty, and iteratively expands the parameter domain until the uncertainty falls below the threshold.

\subsection{Theoretical Justification}

\subsubsection{(a) Uncertainty awareness in approximating Bayesian inversion using diffusion models.}

The reverse diffusion process converges (under standard assumptions) to samples from the true
posterior \(p(\theta \mid y)\). If the training support does not include the true parameter region,
then the learned posterior necessarily exhibits inflated uncertainty:
\[
    \Sigma_\theta \uparrow
    \quad \text{when} \quad \theta^\star \notin \Theta_k.
\]
Thus, the posterior covariance provides a principled diagnostic for missing support.

\subsubsection{(b) Active domain refinement as an optimal-design problem.}
Let \(\Theta_k\) be the current domain. If \(\theta^\star \notin \Theta_k\), the Maximum A Posteriori (MAP) estimate
\[
    \hat{\theta} = \arg\max_{\theta} p_\phi(\theta \mid y^\star)
\]
identifies the nearest region of parameter space where additional training samples will maximally
reduce epistemic uncertainty. 
Expanding \(\Theta_k\) to include \(\hat{\theta}\) minimizes the discrepancy
\[
    \mathrm{KL}\!\left( p(\theta \mid y^\star) \, \big\| \, p_\phi(\theta \mid y^\star) \right),
\]
where $\mathrm{KL}(.)$ denotes the Kullback-Leibler divergence.

Consequently, the sequence of expanded domains
\[
    \Theta_0 \subset \Theta_1 \subset \Theta_2 \subset \cdots
\]
eventually satisfies
\[
    \theta^\star \in \Theta_n \quad \text{for some finite } n,
\]
after which the diffusion posterior contracts around the true parameter region.
\subsection{Practical Limitations of Posterior-Covariance Diagnostics}
In practice, the posterior variance generated by the diffusion-based inverse model is not always a reliable indicator of model misspecification~\cite{nalisnick2018deep}. 
Occasionally, when the true parameters lie outside the training domain, the DM may extrapolate overconfidently and produce posterior distributions with narrow variance.
This is well-known in expressive generative models, which can assign high confidence to regions that are unsupported by training data~\cite{zhang2021understanding}.
Therefore, uncertainty estimated from a single DM may fail to signal the inferred parameters being inconsistent with the training domain, resulting
in premature convergence.
This practical limitation motivates the ensemble-based inference strategy described next, where disagreement across independently trained models is used as a more robust signal of missing training support.

\subsection{Ensemble-based Inference}
To address this issue, an ensemble-based strategy can be used, where multiple DMs are trained independently, each with different random initializations.
The key observation is that, when these DMs are properly trained, if the true parameters lie within the training region, these models converge to similar conditional distributions, due to the fact that the learned inverse mappings are well constrained by data. 
In contrast, when the true parameters are outside the training domain, the inverse problem becomes underconstrained and the learned extrapolations depend sensitively on initialization and training noise.
As a result, the ensemble DMs yield systematic disagreement, even though individual ones may predict with low posterior variance.

Let $\{p_{\phi_k}(\theta|y)\}_{k=1}^K$ denote an ensemble of independently trained DMs. 
The ensemble disagreement can be quantified by the variance of posterior means,
\[
    \mathrm{Var}_k (\mathbb{E}_{p_{\phi_k}} [\theta|y]).
\]
A small variance indicates that parameter inference is well supported by the training data, whereas a large variance suggests extrapolation beyond the training domain. 
A more informative measure of ensemble disagreement compares the predicted posterior distributions directly, for example, by computing pairwise distributional similarities or divergences.

Incorporating ensemble disagreement into the active learning loop provides a more robust criterion for domain refinement than simply posterior variance.
When the ensemble predictions diverge, new parameter samples are generated in the regions suggested by the ensemble outputs, forward model simulations are carried out, and the DMs are retrained.
The active learning process continues until ensemble agreement is reached.

\begin{figure*}[t]
\centering
\includegraphics[width=1.65\columnwidth]{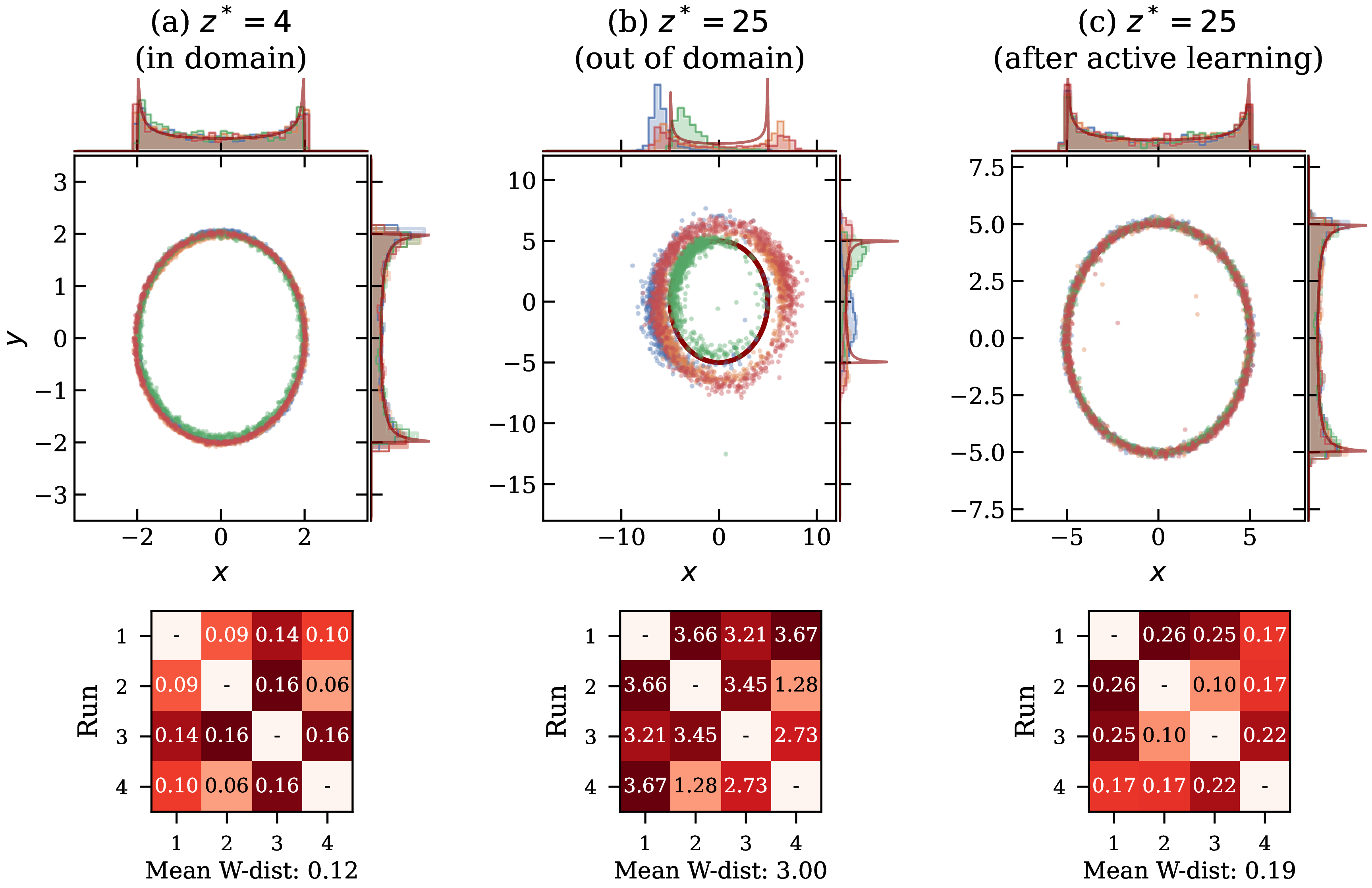}
\caption{Ensemble results (4 independent runs) for the toy inverse problem. Top: posterior samples with true solution (solid circle) and marginal distributions. Bottom: pairwise Wasserstein distance matrices between runs (lower values indicate better agreement). (a) In-domain: consistent predictions with low pairwise Wasserstein distances ($\bar{W} = 0.12$). (b) Out-of-domain: inconsistent predictions with high pairwise Wasserstein distances ($\bar{W} = 3.00$). (c) After active learning: predictions converge with low pairwise Wasserstein distances ($\bar{W} = 0.19$).}
\label{fig:circle}
\end{figure*}
\section{Results}
\subsection{A Toy Problem}
\label{toyproblem}
Let $\theta = (x, y) \in \mathbb{R}^2$ denote unknown parameters and $z \in \mathbb{R}$ the observable, related through the forward model
\begin{equation}
F: \theta \mapsto z = x^2 + y^2.
\label{eq:circle}
\end{equation}
Given an observable $z^*$, the inverse problem admits infinitely many solutions lying on a circle of radius $\sqrt{z^*}$ centered at the origin. This toy inverse problem is ill-posed, making it well-suited for evaluating whether the DM can correctly capture the full posterior distribution.

The DM consists of a 4-layer MLP with hidden dimension 256, conditioned on the observable $z$ through a separate embedding network. We use $T = 100$ diffusion timesteps with a linear noise schedule ranging from $\beta_1 = 10^{-4}$ to $\beta_T = 0.02$. The initial training domain is $\Theta_0 = \{(x, y) : x^2 + y^2 \leq 4\}$, corresponding to radii $r \in [0, 2]$. We generate 50,000 training samples by uniformly sampling parameters from $\Theta_0$ and computing the corresponding observables via~\eqref{eq:circle}, with Gaussian noise $\epsilon \sim \mathcal{N}(0, 0.01)$ added to simulate measurement uncertainty. The model is trained for 20,000 epochs using Adam optimizer~\cite{kingma2017adammethodstochasticoptimization} with learning rate $10^{-4}$ and batch size 256.

After training, we perform inference for both in-domain ($z^* = 4$, corresponding to $r = 2$) and out-of-domain ($z^* = 25$, corresponding to $r = 5$) observables. For the out-of-domain case, we apply the active learning loop described in Section~\ref{activelearningloop}, training until convergence is reached. To assess robustness, we repeat the entire procedure by training 4 DMs with independent random initializations. 

% We evaluate the quality of posterior samples using the reduced chi-square statistic $\chi^2/\text{dof}$, which measures the goodness-of-fit between the sample distribution and the theoretical density. Values close to 1 indicate good agreement.

Figure~\ref{fig:circle} summarizes the results. In case (a), the observable $z^* = 4$ lies within the training range. All four independently trained models consistently recover the posterior distribution, producing samples uniformly distributed along the target circle of radius 2. The resulting marginal distributions closely match the theoretical densities, and the pairwise Wasserstein distances between model predictions are small (mean $\bar{W} = 0.12$), indicating strong agreement across the ensemble.
In contrast, case (b) corresponds to an out-of-domain observable, $z^* = 25$. Here, the independently trained models yield inconsistent posterior predictions scattered across parameter space. This behavior is reflected by substantially larger pairwise Wasserstein distances (mean $\bar{W} = 3.00$), signaling ensemble disagreement and model extrapolation beyond the training support.
Case (c) shows the inference results for $z^* = 25$ after training with active domain expansion. Following the active learning loop, all four models recover the correct posterior, with samples concentrated along the target circle of radius 5. The pairwise Wasserstein distances decrease remarkably (mean $\bar{W} = 0.19$), returning to levels comparable with the in-domain case. These results demonstrate that ensemble disagreement, quantified by pairwise Wasserstein distance, provides a reliable indicator of out-of-domain extrapolation. Then, the active learning loop systematically expands the training domain toward regions supported by the observations, enabling the DM to recover the correct posterior distribution.
Here in Figure~\ref{fig:circle}, we use $K=4$ independently trained models in the experiments to balance visualization clarity and trigger robustness. In practice, larger ensembles can improve reliability at higher computational cost.

\subsection{Parameterization of Particle Momentum Distribution}
We use a simulated inclusive Deep Inelastic Scattering (DIS) setup, in which the Particle Momentum Distribution (PMD) is parameterized as a quark correlation function (QCF), to demonstrate the effectiveness of the active diffusion-based inverse solver in a practical application. 
In inclusive DIS experiments, high-energy leptons scatter off nuclear targets, producing complex final-state particle showers. 
By measuring the kinematics of the scattered lepton, one can extract information about the internal quark and gluon structure of the target nucleus.

The PMD encodes the probability density of observing particles with specific momenta following the lepton–nucleus interaction. 
As such, it provides an indirect representation of the underlying QCFs that characterize the momentum and spatial correlations of quarks and gluons. Inferring QCFs from measured PMDs constitutes a challenging inverse problem that is commonly addressed using Bayesian inference methods with physically motivated priors \cite{bishop2006pattern}.

The dimensionality of the PMD is directly determined by that of the underlying three-dimensional QCFs, which in principle requires at least three final-state particle measurements to fully constrain the distribution. 
Traditional histogram-based reconstruction methods are widely used to estimate PMDs, but these approaches become increasingly inadequate in higher dimensions, where they tend to obscure important multi-particle correlations. 
To illustrate the key ideas of our framework in a controlled setting, we therefore perform a simplified proxy calculation that captures essential features of realistic QCD-based models used in global analyses \cite{cocuzza2022polarized}.

The PMDs for a simplified version of DIS on protons and neutrons are given by
% \begin{subequations}
\begin{align}
\begin{split}
    \pmb{\sigma}_1(x;p)=4u(x;p) + d(x;p), \\
    \pmb{\sigma}_2(x;p)=u(x;p) + 4d(x;p)\,.
    \end{split}
\label{eq:cross}
\end{align}
% \end{subequations}
where $\pmb{\sigma}_1$ and $\pmb{\sigma}_2$ are cross sections (un-normalized probability distributions) and $u(x)$ and $d(x)$ are the universal 1D QCFs called up- and down-quark PDFs, respectively, which are weighted by their charge squared in the PMDs. QCFs for the proxy problem with two channels are defined by:

\begin{align}
\begin{split}
    u(x;p)=&N_u x^{a_u} (1-x)^{b_u}\\
    d(x;p)=&N_d x^{a_d} (1-x)^{b_d}\,
\end{split}
\label{eq:toy: problem}
\end{align}
\noindent where $x \in (0,1)$ and $\{N_u,a_u, b_u, N_d,a_d, b_d\}$ is the unknown parameter vector to be determined. 
We collect events $\{\sigma_p^{o},\sigma_n^o\}$ generated by model~\eqref{eq:toy: problem} and filtered through cross-sections defined in~\eqref{eq:cross} for specific ranges of the shape parameters $\{N_u,a_u, b_u, N_d,a_d, b_d\}$.
From these QCFs, we create PMDs and then sample the PMDs to generate the physics events as the observables.
\begin{figure}[t]
    \centering
    \includegraphics[width=0.90\columnwidth]{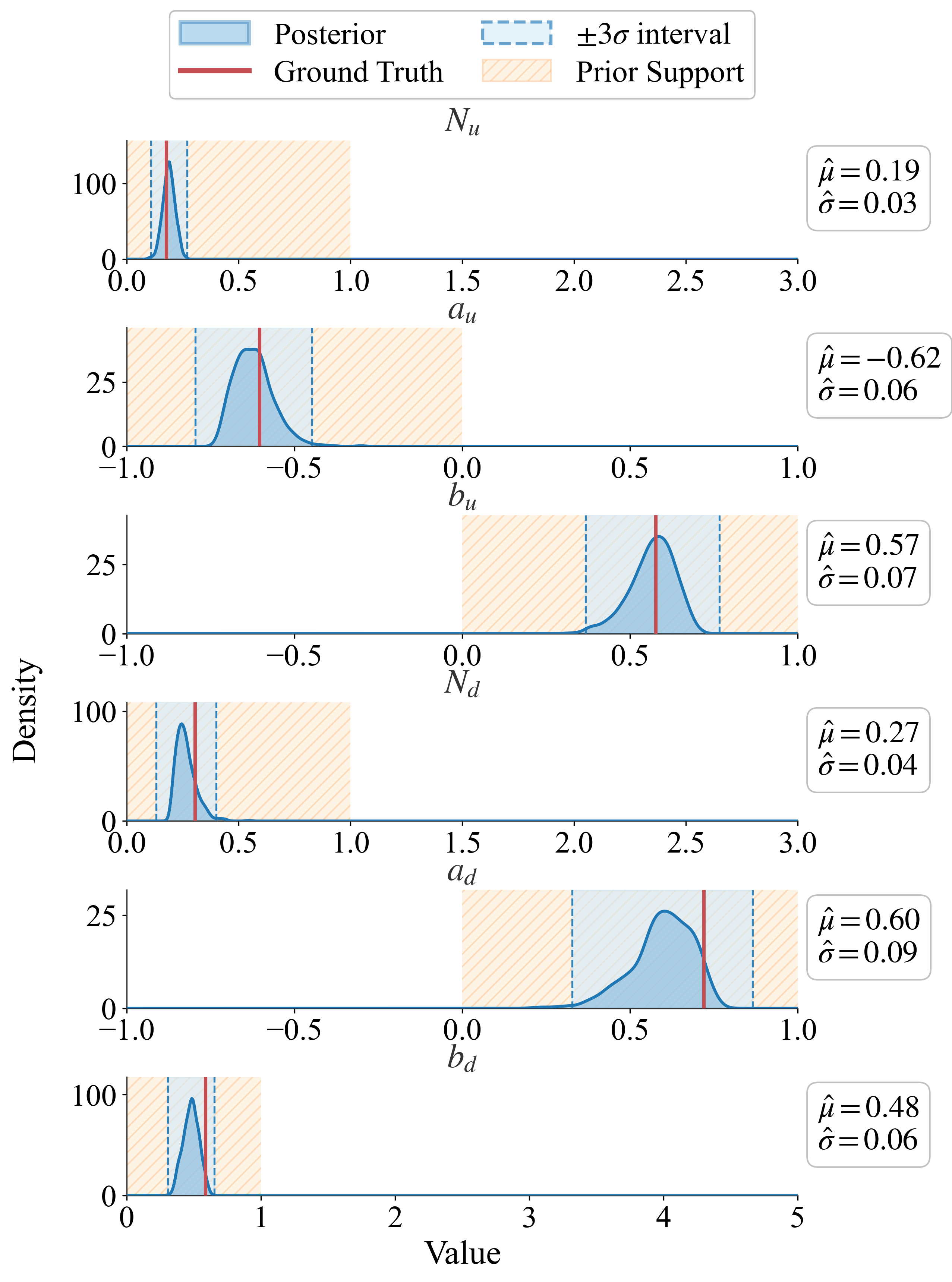}
    \caption{In-domain inference with true parameters in the training domain. The model accurately recovers all parameters with low uncertainty. The shaded orange region denotes the prior support.}
    \label{fig:qcd-indomain}
\end{figure}

\begin{figure}[t]
    \centering
    \includegraphics[width=0.90\columnwidth]{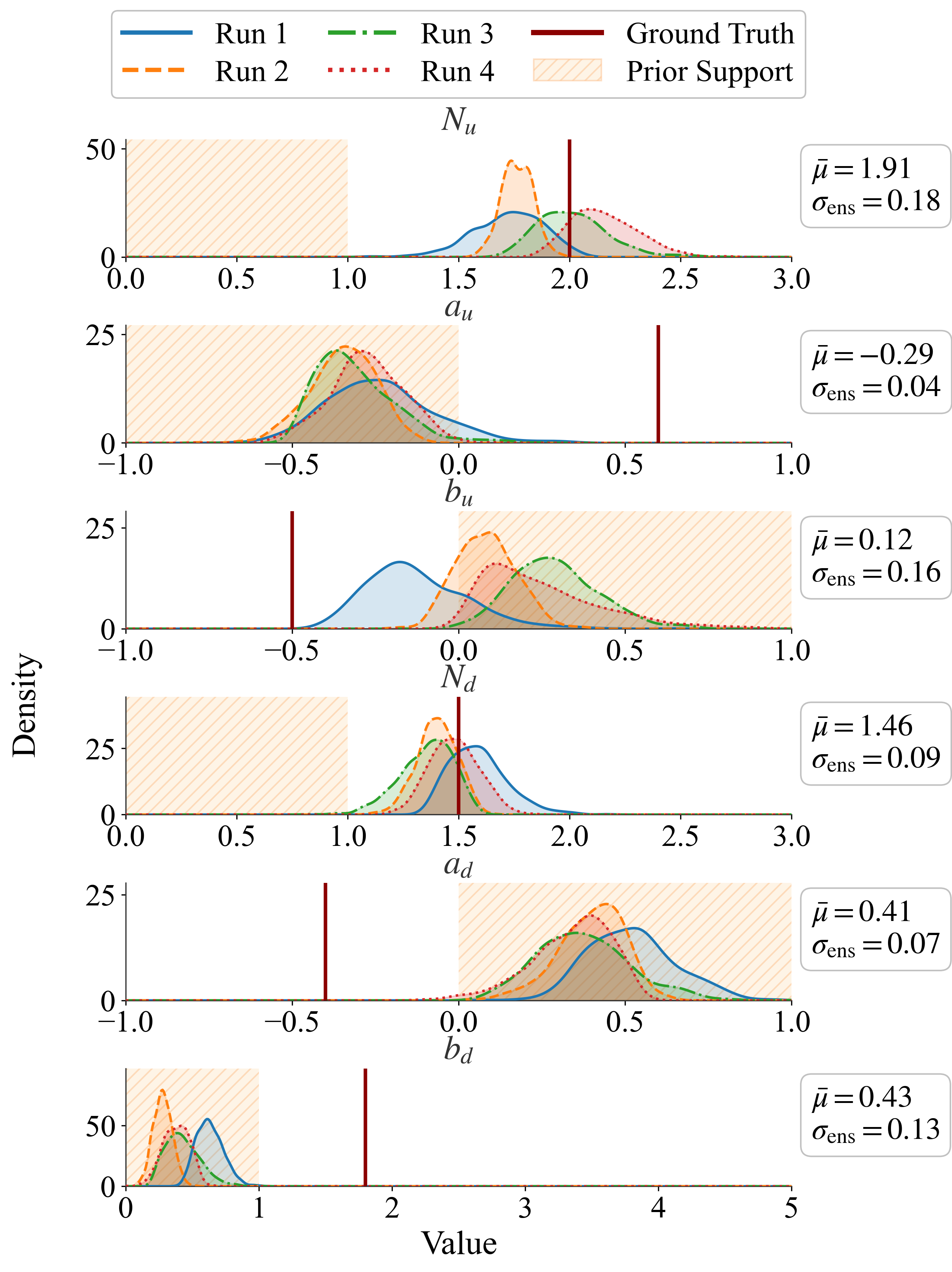}
    \caption{Out-of-domain inference without active learning. Four independently trained models (shown in different colors) produce inconsistent predictions with high ensemble disagreement ($\sigma_{ens}$), indicating model misspecification.}
    \label{fig:qcd-ood-noal}
\end{figure}

\begin{figure}[t]
    \centering
    \includegraphics[width=0.90\columnwidth]{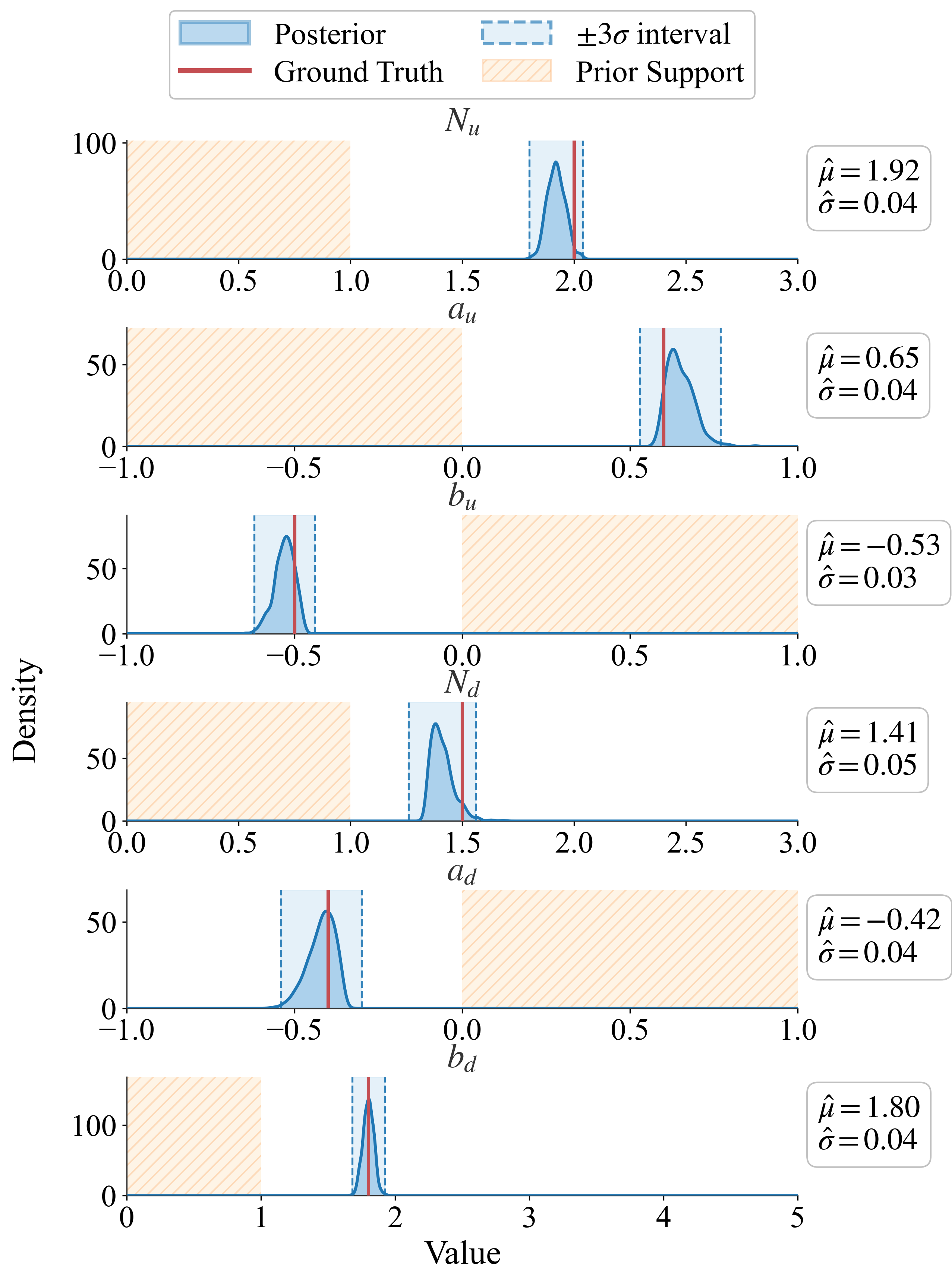}
    \caption{Out-of-domain inference with active learning. After adaptive domain expansion, the model successfully recovers all parameters, where all ground-truth values are within three standard deviations of the corresponding posterior means.}
    \label{fig:qcd-ood-al}
\end{figure}

\subsection{QCD Parameter Inference Results}
We implement the active diffusion-based inverse solver using a PointNet-Transformer architecture~\cite{qi2017pointnet,vaswani2017attention} as the backbone network, which processes event-level observables through permutation-invariant encoding followed by self-attention layers. The model is trained using an Adam optimizer with learning rate $10^{-4}$ and batch size 64, where each parameter sample is conditioned on 10,000 events generated through the forward model described by Equations~\eqref{eq:cross} and~\eqref{eq:toy: problem}. The diffusion process uses $T = 100$ timesteps with a linear noise schedule from $\beta_1 = 10^{-4}$ to $\beta_T = 0.02$. The initial parameter bounds are $N_u, N_d \in [0, 1]$, $a_u \in [-1, 0]$, and $a_d, b_u, b_d \in [0, 1]$.
\paragraph{In-domain Inference.}
When ground truth parameters are sampled from the interior of the prior distribution, the trained DM predicts all six QCD parameters with high fidelity (Figure~\ref{fig:qcd-indomain}). The posteriors concentrate tightly around true values with small standard deviations ranging from $0.03$ to $0.06$, confirming reliable inference in the training support.

\paragraph{Out-of-domain Inference without Active Learning.}
We construct a challenging out-of-domain test case in which all true parameters lie outside the training bounds: $N_u = 2.0$, $a_u = 0.6$, $b_u = -0.5$, $N_d = 1.5$, $a_d = -0.4$, and $b_d = 1.8$. An ensemble of four independently initialized DMs produces the failed inference results shown in Figure~\ref{fig:qcd-ood-noal}. For each model in the ensemble, the inferred posterior distributions exhibit substantially larger standard deviations than those observed in in-domain inference, indicating increased uncertainty and weakened posterior concentration.

More importantly, the four independently trained DMs yield inconsistent posterior estimates. The ensemble disagreement $\sigma_{\text{ens}}$, defined as the standard deviation of posterior means across models, is significantly elevated for these parameters, ranging from $0.04$ to $0.18$. The lack of consensus reveals that the predictions are not supported by the training data distribution, serving as a clear and practical indicator of out-of-domain extrapolation. Overall, these results demonstrate that ensemble disagreement provides a practical and reliable signal for detecting model misspecification.
% We construct a challenging test case where all parameters lie outside the training bounds: $N_u = 2.0$, $a_u = 0.6$, $b_u = -0.5$, $N_d = 1.5$, $a_d = -0.4$, and $b_d = 1.8$. An ensemble of four independently initialized models reveals the failed predictions shown in Figure~\ref{fig:qcd-ood-noal}. One can find that, for each independently trained model, the predictions of the parameters yield significantly wider standard deviations than the cases in in-domain inference. Moreover, the ensemble disagreement $\sigma_{\text{ens}}$, measured as the standard deviation of posterior means across the four models, is high for these parameters, ranging from $0.04$ to $0.18$. This indicates that the four independently trained diffusion models disagree with each other, suggesting out-of-domain extrapolation is detected.

% For $b_u$, the ensemble disagreement is large ($\sigma_{\text{ens}} = 0.16$), indicating substantial variability across models, even though the individual posteriors remain relatively broad. The ensemble mean for $N_u$ ($\bar{\mu} = 1.91$) is close to its true value 2.0, but $a_d$ shows $\bar{\mu} = 0.41$ versus the true $-0.4$, and $b_d$ gives $\bar{\mu} = 0.43$ versus 1.8. The large variability of posterior means across independently trained models indicates that the inferred parameters are not well constrained by the training data. This confirms that ensemble disagreement can detect out-of-domain extrapolation.
\paragraph{Out-of-domain Inference with Active Learning.}
We apply the active diffusion-based inference to the same out-of-domain test case by iteratively expanding the parameter domain through active learning. At each iteration, the DM generates candidate parameter samples via reverse diffusion conditioned on the observed events. These event samples identify regions of parameter space associated with high posterior uncertainty or ensemble disagreement. Additional training data are then generated in these regions using the forward model by Equations~\eqref{eq:cross} and~\eqref{eq:toy: problem}, and the DM is refined on the augmented training set. This process is repeated until convergence is reached.

Figure~\ref{fig:qcd-ood-al} summarizes the inference results after convergence of the active learning loop. The posterior means are estimated as $\hat{N_u} = 1.92$, $\hat{a_u} = 0.65$, $\hat{b_u} = -0.53$, $\hat{N_d} = 1.41$, $\hat{a_d} = -0.42$, and $\hat{b_d} = 1.80$. The inferred posterior distributions accurately recover the ground-truth parameters, with all ground-truth values lying within three standard deviations of the corresponding posterior means. Moreover, posterior uncertainties contract to levels comparable to in-domain inference ($\hat{\sigma} \approx 0.03$–$0.05$), despite all inferred posteriors lying entirely outside the original training support. These results demonstrate that the active diffusion framework can reliably detect and correct out-of-domain extrapolation, autonomously discovering relevant parameter regions and enabling accurate inverse inference without prior knowledge of the valid parameter domains.

\section{Discussion}
\subsection{Relation to Shooting Method}
Our method is conceptually close to the classical shooting method in classical numerical analysis~\cite{stoer1980introduction}.
In shooting methods for boundary-value problems, the initial conditions, typically treated as unknown parameters, are iteratively adjusted so that the forward solution of a differential equation can gradually satisfy the boundary condition, a prescribed terminal constraint. 
Each iteration includes executing the forward model, evaluating the mismatch at the boundary, and updating the parameter estimate accordingly.
It is important to note that the shooting method does not require prior knowledge of the domain of the parameter values; instead, it progressively refines them based on the residual error as feedback from the forward model.

In the inverse problem, the unknown parameters play a similar role as the unknown initial conditions in the shooting methods.
The DM serves as a surrogate inverse mapper, proposing candidate parameter solutions that are consistent with the observables.
When the true parameters lie outside the initial training domain, the diffusion posterior exhibits large uncertainty, indicating a mismatch with respect to the observables. 
The active learning loop can therefore be interpreted as a probabilistic generalization of the shooting methods. 
Instead of updating the estimate of a single parameter, the support of the parameter distribution is updated by selectively expanding the training domain in new regions suggested by the DM. 
The forward model is then carried out on these candidate parameters, providing new parameter-observable mapping information that further improves the inverse surrogate.

Analogue to the shooting methods, the evaluations of the forward models are often computationally costly and need to be chosen smartly to maximize information gain.
A key distinction between this method and the shooting methods is that shooting methods use deterministic residuals to guide parameter updates, whereas the DM provides a full posterior distribution over parameters conditioned on the observables. 
From this point of view, the active learning DM may be considered as a shooting method in distribution space.
The iterative expansion of the parameter domain replaces the iterative correction of initial conditions, and the convergence is reached when the posterior distribution contracts to a stable region consistent with the observables.

\subsection{Ensemble-based Analysis}
The ensemble disagreement is similar to residual sensitivity in shooting methods. 
Different initial guesses leading to divergent terminal behavior indicate that either the boundary value problem is ill-conditioned or no valid solutions in the current search region.
Similarly, disagreement among trained DMs ensemble suggests that the parameter domain lacks sufficient coverage.
Using the measure of ensemble disagreement to trigger active learning allows the algorithm to identify the regions of parameter space where additional forward-model evaluations are most likely informative toward the true parameters.  

\subsection{Comparison with Other Generative Models}
\begin{table}[t]
\centering
\footnotesize
\begin{tabular}{lccc}
\toprule
\textbf{Method} & \textbf{In-domain} & \textbf{OOD fixed} & \textbf{OOD +AL} \\
\midrule
MDN & 0.067 & 2.497 & 0.082 \\
INN & 0.003 & 1.679 & 0.025 \\
NF & 0.045 & 107.243 & 0.079 \\
\textbf{DM} & 0.015 & 1.279 & 0.032 \\
\bottomrule
\end{tabular}
\caption{Comparison of various generative models on the toy circle inverse problem.}
\label{tab:baseline}
\end{table}

In addition to DM, other generative AI models can also be incorporated into the active learning inverse inference framework under incomplete priors.
For completeness, we compare against other learned inverse solvers, including MDN, INN, and RealNVP-style NF, on the same toy problem specified in Section~\ref{toyproblem}. Table~\ref{tab:baseline} reports the mean radial error (MRE), where all methods are trained on $z \in [0,4]$ and tested in-domain at $z^*=4$ and out-of-domain (OOD) at $z^*=25$.
One can find that all fixed-domain methods perform well within the training domain but degrade substantially under OOD conditions, as reflected by large MRE values. When augmented with the proposed active learning framework (``+AL''), all methods recover near in-domain accuracy. This result demonstrates that the proposed framework is model-agnostic and can be effectively integrated with a broad class of generative models for inverse problems, rather than being specific to DMs.

\subsection{Computational Cost and Practical Considerations}
The main computational cost of the active framework comes from additional forward-model evaluations in expanded parameter regions and from retraining or fine-tuning the generative model. Particularly, in the QCD proxy experiment, each training step requires approximately 2.6 seconds for forward simulation with 64 parameter samples and 10,000 events per sample, while the DM training update takes approximately 0.03 seconds per step.
More generally, the overall computational efficiency is inherently application-dependent. It is primarily governed by the cost of the forward simulation, the distance from the true parameters to the initially covered training domain, the dimensionality of the parameter space, and the geometric complexity of the parameter landscape.

\section{Conclusion}
In this work, we introduce an active diffusion-based framework for solving inverse problems under incomplete prior knowledge of the parameter domain.
By learning the probabilistic mapping between parameters and observables, the DM achieves inverse inference as conditional generation, addressing nonlinearity and ill-posedness in inverse problems.
Without relying on fixed parameter bounds, the active learning loop enables adaptively discovering and refining the relevant regions of parameter space, ensuring reliable inference even when the true parameters are not included in the initial training domain.
The active diffusion-based inference is validated using a toy inverse problem with infinitely many solutions and a QCF parameterization problem in Quantum Chromodynamics analysis of nucleon structure.
The posterior distributions are correctly recovered and the out-of-domain failures are effectively corrected through active retraining.
It is important to note that when a sufficiently large valid parameter domain is known in advance and sampling this domain is computationally affordable, fixed-domain training is sufficient. Our setting addresses the complementary case in which the valid parameter domain is unknown, effectively unbounded, or too high-dimensional to cover uniformly.

Several promising research directions emerge from this work.
In the future, we will focus on scaling the active diffusion-based model to higher dimensional parameter spaces and more sophisticated forward models.
In particular, we will investigate surrogate approximated forward modeling and physics-informed constraints to achieve better efficiency and stability.
Moreover, we will apply our methods to real experimental data where data uncertainties, systematic uncertainties, and model uncertainties coexist.
This will be a critical step toward building active generative inference as a practical tool with uncertainty quantification for scientific discovery in large-scale inverse problems.

\section*{Acknowledgments}
This work is partially supported by the U.S. Department of Energy, Office of Science, Office of Nuclear Physics, Office of Advanced Scientific Computing Research through the Scientific Discovery through Advanced Computing (SciDAC) program, under contracts DE-AC02-06CH11357, DE-AC05-06OR23177, and DE-SC0023472, for the award {\it Femtoscale Imaging of Nuclei using Exascale Platforms}, the EXCLAIM collaboration under the DOE grants DE-SC0016286 and DE-SC0024644, and by the Center for Nuclear Femtography (CNF), administrated by the Southeastern Universities Research Association under an appropriation from the Commonwealth of Virginia under contract No. C2024-FEMT-011-02.


\bibliographystyle{named}
\bibliography{ijcai26}

\end{document}